\documentclass[runningheads]{llncs}
\usepackage{graphicx}

\usepackage{tikz}
\usepackage{comment}
\usepackage{amsmath,amssymb} 
\usepackage{color}

\usepackage[accsupp]{axessibility}  

\usepackage{booktabs}
\usepackage{bm}
\usepackage{float}

\begin{document}
\pagestyle{headings}
\mainmatter
\def\ECCVSubNumber{2802}  

\title{Resolving Copycat Problems in Visual Imitation Learning via Residual Action Prediction} 

\titlerunning{ECCV-22 submission ID \ECCVSubNumber} 
\authorrunning{ECCV-22 submission ID \ECCVSubNumber} 
\author{Anonymous ECCV submission}
\institute{Paper ID \ECCVSubNumber}

\titlerunning{Resolving Copycat Problems via Residual Action Prediction}
%
\author{Chia-Chi Chuang$^{*}$\inst{1}
\and Donglin Yang$^{*}$\inst{1}
\and Chuan Wen$^{*}$\inst{1}
\and Yang Gao$^{\dag}$\inst{1,2}}
\authorrunning{Chia-Chi et al.}
%
\institute{Institute for Interdisciplinary Information Sciences, Tsinghua
University \\
\and Shanghai Qi Zhi Institute \\
\email{\{zhuangjq19,ydl18,cwen20\}@mails.tsinghua.edu.cn}, \email{gaoyangiiis@tsinghua.edu.cn} }
\maketitle

\begin{abstract}

Imitation learning is a widely used policy learning method that enables intelligent agents to acquire complex skills from expert demonstrations. The input to the imitation learning algorithm is usually composed of both the current observation and historical observations since the most recent observation might not contain enough information. This is especially the case with image observations, where a single image only includes one view of the scene, and it suffers from a lack of motion information and object occlusions. In theory, providing multiple observations to the imitation learning agent will lead to better performance. However, surprisingly people find that sometimes imitation from observation histories performs worse than imitation from the most recent observation. In this paper, we explain this phenomenon from the information flow within the neural network perspective. We also propose a novel imitation learning neural network architecture that does not suffer from this issue by design. Furthermore, our method scales to high-dimensional image observations. Finally, we benchmark our approach on two widely used simulators, CARLA and MuJoCo, and it successfully alleviates the copycat problem and surpasses the existing solutions.
\end{abstract}

\let\thefootnote\relax\footnotetext{$^{*}$ Equal contribution.}
\let\thefootnote\relax\footnotetext{$^{\dag}$ Corresponding author.}

\section{Introduction}

Learning to control in complex environments is a challenging task. Imitation learning is a powerful technique that learns useful skills from a pre-collected expert demonstration~\cite{widrow1964pattern,argall2009survey,Ross2011,osa2018algorithmic,ho2016gail}. Compared with reinforcement learning~\cite{mnih2015human,schulman2015trust,schulman2017proximal}, imitation learning is generally more data-efficient and does not require destructive exploration. Behavioral cloning (BC)~\cite{pomerleau1989alvinn,muller2006off,mulling2013learning,giusti2015machine,bojarski2016end} is one of the most widely used imitation algorithms. It directly mimics the expert behavior by learning the mapping from the observations to the expert actions with supervised learning. Imitation learning has achieved many successes in the past in domains like autonomous driving~\cite{pomerleau1989alvinn,bojarski2016end} and drone flying~\cite{loquercio2019deep}. Imitation learning has also become an essential component in many other policy learning algorithms~\cite{Ross2011,levine2013guided}.

In an Markov Decision Process (MDP), the input to the behavioral cloning algorithm can be the current state (Behavioral Cloning from Single Observation, i.e., BCSO). However, in practice, the agent's observation might be far from Markovian. This is the case, especially with visual observations. The most recent visual frame usually misses essential information, such as the objects' motion and appearance that are occluded in the current frame but visible in previous frames. Thus in many applications, it is more reasonable to provide the behavioral cloning algorithm with the observation histories (Behavioral Cloning from Observation Histories, i.e., BCOH) instead of a single observation. This allows the agent to access the state of the problem better.

\begin{figure}[t]
    \setlength{\belowcaptionskip}{-0.5cm}
    \centering
    \includegraphics[width=0.8\textwidth]{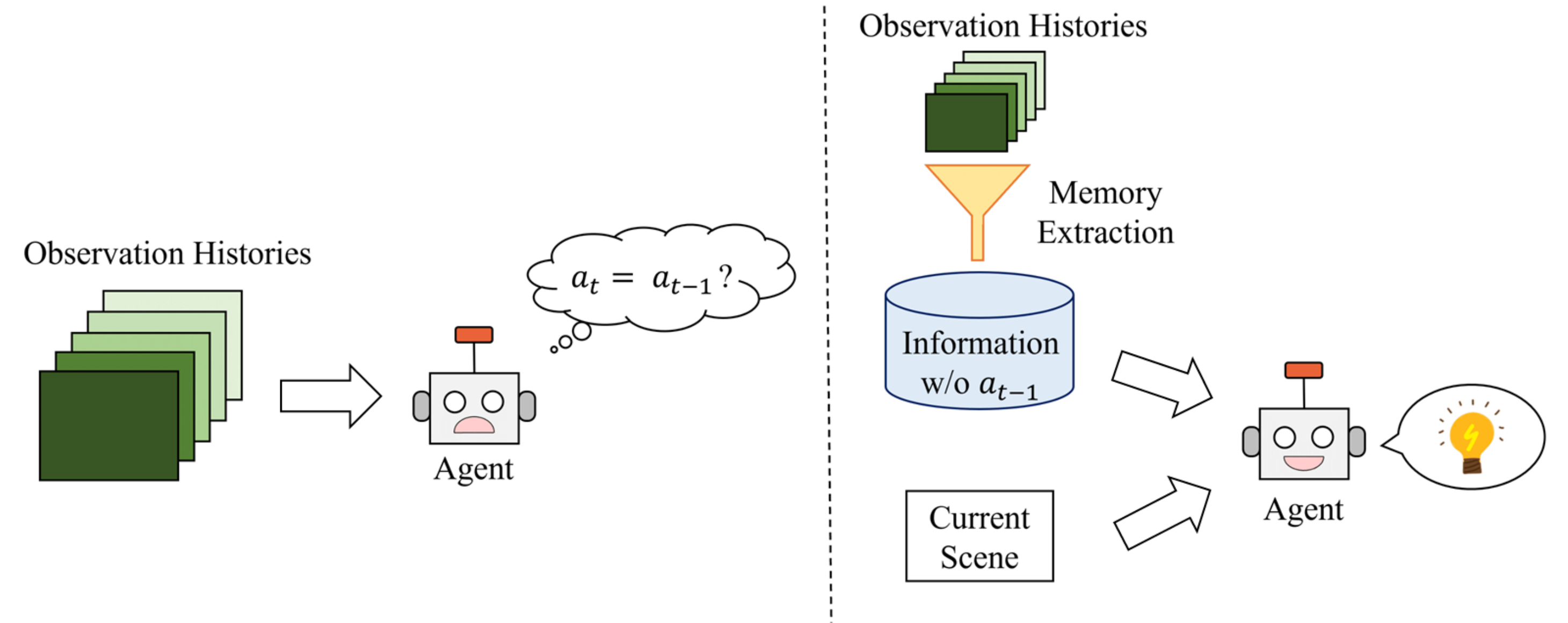}
    \caption{\textbf{The copycat problem \& our solution:} Left: Behavioral cloning from observation histories might learn a shortcut that directly outputs its previous action as the current action. Unfortunately, this leads to inferior performance during the test time. Right: We propose to solve the problem by a copycat-free memory extraction module. The history information is carefully extracted such that the shortcut no longer exists.}
    \label{fig:concept}
\end{figure}

However, recent works~\cite{dehaan2019causal,wang2019monocular} find that behavioral cloning with observation histories can sometimes perform even worse than behavioral cloning from the most recent observation. This phenomenon is counter-intuitive: the more complete input features (observation history) should lead to better downstream performance, but in practice, it won't! \cite{dehaan2019causal,chuan2020fighting,wen2021keyframefocused,wen2022fighting} explain this phenomenon as BCOH is prone to learn an action prediction shortcut instead of the correct concept. BCOH predicts the current action $a_t$ from the observation histories $o_{t}, o_{t-1}, \cdots$. However, the observation histories are actually obtained by executing previous expert actions $a_{t-1}, a_{t-2}, \cdots$ in the environments. Usually, the actions of an agent transit smoothly. Thus the BCOH agent tends to recover $a_{t-1}$ from the observation histories and predict $a_t$ based on that. This phenomenon is referred to as \texttt{copycat problem} or \texttt{inertia problem} in the literature~\cite{chuan2020fighting,wen2021keyframefocused,codevilla2019exploring}. It is a kind of causal confusion that severely limits the application of behavioral cloning algorithms to a broader range of the problem. 

Early works~\cite{dehaan2019causal,chuan2020fighting,wen2021keyframefocused} have proposed several approaches to resolve the causal confusion problem. However, they are either limited to only dealing with low-dimensional state-based environments~\cite{chuan2020fighting} or suffer from performance limitations when dealing with high-dimensional image observations~\cite{dehaan2019causal,wen2021keyframefocused}. In this paper, we approach the problem from a neural network information flow view and propose a neural network architecture that does not have the copycat problem by design. Since the BCOH method learns the incorrect solution because the previous action $a_{t-1}$ is a shortcut solution, we propose to design the neural network architecture such that it cannot learn to act based on the previous actions. Fig.~\ref{fig:concept} illustrates the concept of why copycat problem occurs and our solution. Unlike previous approaches, which achieve similar effects by using adversarial training~\cite{chuan2020fighting} or using per sample re-weighting~\cite{wen2021keyframefocused}, our policy avoids the incorrect solution more thoroughly by cutting the information flow of the wrong clue. Thus, our method trains more easily and achieves better task performances. 

We benchmark our method in a challenging autonomous driving environment, CARLA~\cite{Dosovitskiy17}. CARLA is a visually realistic driving simulator. It is widely used as a benchmark for autonomous driving and imitation learning~\cite{codevilla2018end,heinze2017conditional,codevilla2019exploring}. 
Besides CARLA, we evaluate the performance in a commonly used robotics simulator, MuJoCo~\cite{todorov2012mujoco}, with image-based observation. Our method outperforms all previous approaches on these challenging benchmarks. 

\section{Related Work}

\noindent \textbf{Imitation Learning and Copycat Problems:~~}
Imitation learning~\cite{widrow1964pattern,argall2009survey,osa2018algorithmic} is a powerful technology to learn complex policies from expert demonstrations. Among the different imitation learning methods, behavioral cloning is a simple but effective paradigm that directly regresses the expert actions from the observations. However, like other imitation methods, behavioral cloning suffers from the distributional shift that small errors will accumulate over time during testing and finally make the imperfect imitators encounter out-of-distribution states~\cite{Ross2011}. We focus on a specific phenomenon arising under the distributional shift in the partially observed settings -- the copycat problem~\cite{chuan2020fighting,wen2021keyframefocused,wen2022fighting}.
Although environmental interactions~\cite{ho2016gail,dehaan2019causal,Brantley2020} or a queryable expert~\cite{Ross2011,Sun2017,laskey2017dart,sun2018truncated,spencer2021feedback} can resolve it, purely offline methods~\cite{chuan2020fighting,bansal2018chauffeurnet,wen2021keyframefocused,wen2022fighting} are understudied. The specific definition and existing solutions for the copycat problem will be introduced in detail in Sec.~\ref{sec:copycat}. In this paper, we propose a new neural network architecture for imitation policies to solve the copycat problem more thoroughly.

\noindent \textbf{Shortcut Learning:~~}
While the numerous success stories of deep neural networks (DNN) have rapidly spread over science, industry, and society, its limitations are coming into focus. For example, in computer vision, DNN image classifier tends to rely on texture rather than shape~\cite{geirhos2018imagenet} and the background rather than objects~\cite{beery2018recognition,wen2022fighting}. And in NLP, some language models turn out to depend on spurious features like word length without understanding the content of a sentence~\cite{niven2019probing,mccoy2019right}. \cite{geirhos2020shortcut} summarize this phenomenon and name it ``shortcut learning," i.e., the DNNs prefer to learn the more straightforward solution (shortcut) rather than taking more effort to understand the intended solution. We regard the copycat problem as an instance of shortcut learning: when the scenes and labels change smoothly, the neural networks tend to cheat by extrapolating from the historical information, a shortcut solution. Beyond imitation learning, the copycat problem also exists in other sequential tasks, such as robotics manipulation~\cite{ajay2021matters}, tracking~\cite{zhou2020tracking}, forecasting~\cite{hu2021safe}, etc. Our method may generalize to these tasks, and we leave this to our future work.

\section{Preliminaries}

\subsection{Partially Observed Markov Decision Process}
The decision process of an agent equipped with sensors can be best described by a partially observed Markov decision process (POMDP) since the equipped sensors such as cameras or LIDARs can only perceive part of the environment at each time step. A POMDP can be formalized as a tuple $(\mathbf{S},\mathbf{A},\mathbf{T},r,\mathbf{O})$, where $\mathbf{S}$ is the state space of the environment, $\mathbf{A}$ is the action space of the agent, $\mathbf{T}$ is the transition probabilities between states with a given action, $r:\mathbf{S}\times\mathbf{A}\to\mathbb{R}$ is the reward function and $\mathbf{O}$ is the observation space of the agent. In practice, the dimension of $\mathbf{S}$ is much greater than the dimension of $\mathbf{O}$. At each time step $t$, the agent has no access to the underlying true state $s_{t}$ and has to take the action $a_{t}$ according to the observation $o_{t}$. 
To deal with partial observation, it is common to take the historical information into account~\cite{murphy2000survey,mnih2015human,bansal2018chauffeurnet,chuan2020fighting,wen2021keyframefocused}, constructing observation history $\tilde{o}_{t}=[o_{t},o_{t-1},\cdots,o_{t-H}]$ where $H$ is the history length. 

In the imitation learning setup, we are given a demonstration dataset composed of observation-action tuples: $\{(o_t, a_t)\}$. They represent good behaviors, i.e., the trajectories that achieve high accumulated rewards. However, the reward is not provided in the imitation learning setup. Instead, the goal of imitation learning is to learn a function that maps observation histories $\tilde{o}_t$ to the expert action $a_t$ that leads to high accumulated rewards. 

\subsection{Behavioral Cloning:}
Among all the variants of imitation learning, we focus on behavioral cloning (BC), a straightforward but powerful approach to mimic the expert behaviors from a pre-collected demonstration $\mathcal{D}=\left\{(o_{i},a_{i})\right\}_{i=1}^{N}$, where the $N$ is the number of samples. BC reduces the complex policy learning into supervised learning by maximizing the log-likelihood of the action $a_{t}$ conditioned on the observation history $\tilde{o}_{t}$ so that the agent can behave as similarly to the expert as possible, i.e., $\theta^*=\arg\max_{\theta}\mathbb{E}_{\mathcal{D}}[\log P(a_t|\tilde{o}_{t}; \theta)]$. $H=0$ is BC from single observation (BCSO), and $H>0$ is BC from observational histories (BCOH).

\begin{figure}[t]
    \setlength{\belowcaptionskip}{-0.5cm}
    \centering
    \includegraphics[width=0.6\textwidth]{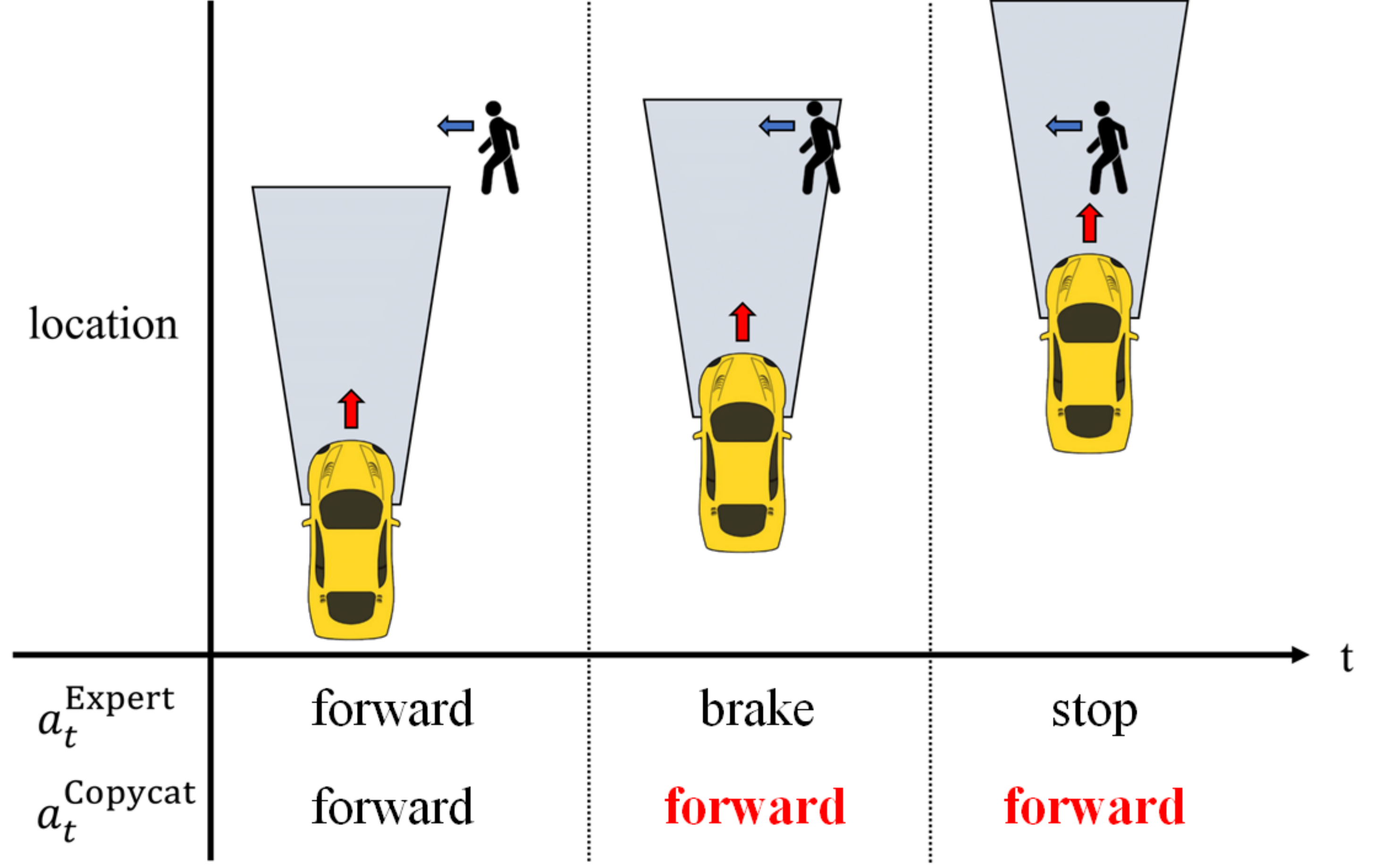}
    \caption{\textbf{An illustrative example of the copycat problem:}
    The autonomous vehicle encounters a pedestrian. The trapezoidal area indicates the region can be observed by the driving agent. The expert agent would brake as soon as it sees the pedestrian, but the copycat agent tends to repeat previous actions and move forward.}
    \label{fig:copycat}
\end{figure}

\subsection{Copycat Problem}
\label{sec:copycat}

The copycat problem is a phenomenon that the BC agent tends to infer a ``copycat'' action similar to the previous action. When the copycat problem happens, both the training and the validation loss are low, but the in-environment evaluation has poor performance. It is not an instance of the over-fitting problem. Prior works \cite{bansal2018chauffeurnet,codevilla2019exploring} showed using only a single observation $o_t$ can achieve better performance than using multiple observations on their tasks. In addition, \cite{dehaan2019causal,chuan2020fighting,wen2021keyframefocused} show that the agent can completely fail when the input contains information about the previous actions. The copycat problem is a widely existing phenomenon in many downstream tasks when using BC, such as autonomous driving~\cite{wen2021keyframefocused,wang2019monocular} and robot control~\cite{dehaan2019causal,chuan2020fighting}. 

Fig.~\ref{fig:copycat} displays a concrete example of the copycat problem. In a driving scenario, the expert brakes immediately when it observes a person crossing the road so it can stop timely to prevent a traffic accident from happening. On the other hand, a copycat agent tends to keep its own previous action and usually brakes late or doesn't brake at all. Similar failure cases have been confirmed by multiple previous works~\cite{wang2019monocular,wen2021keyframefocused}.

The reason why copycat problems happen in BC has been investigated in previous works~\cite{dehaan2019causal,chuan2020fighting,wen2021keyframefocused}. It happens because there exists a shortcut from the input observation history $\tilde{o}_t$ to the target current action $a_t$. The input observation history is obtained by executing the expert actions $a_{t-H}, a_{t-H+1}, \cdots , a_{t-1}$ in the environment. Since most sequential decision problems exhibit a smooth transition property, expert actions also usually change smoothly. This means the previous action $a_{t-1}$ is very close to the prediction target $a_t$. In many cases, the agent predicts the action $a_t$ either completely or partially based on the information of the previous action $a_{t-1}$. However, decisions based on an agent's own previous action could be wrong in many cases, such as the vehicle stopping for the pedestrian case discussed above.

\section{Methodology}
\label{sec:method}

Since the copycat problem is caused by the model learning the spurious prediction pathway from the previous action $a_{t-1}$ to the current action $a_t$, in this section, we propose a neural network architecture that doesn't have this pathway by design. I.e., we would like our neural network to be unable to access the previous action $a_{t-1}$ information according to the neural network architecture. We note that similar ideas have been explored by \cite{chuan2020fighting}. However, they propose to achieve this with adversarial networks, and we show that our solution has a much better scaling property than the previous one.

\begin{figure}[h]
    \setlength{\belowcaptionskip}{-0.5cm}
    \centering
    \includegraphics[width=0.6\textwidth]{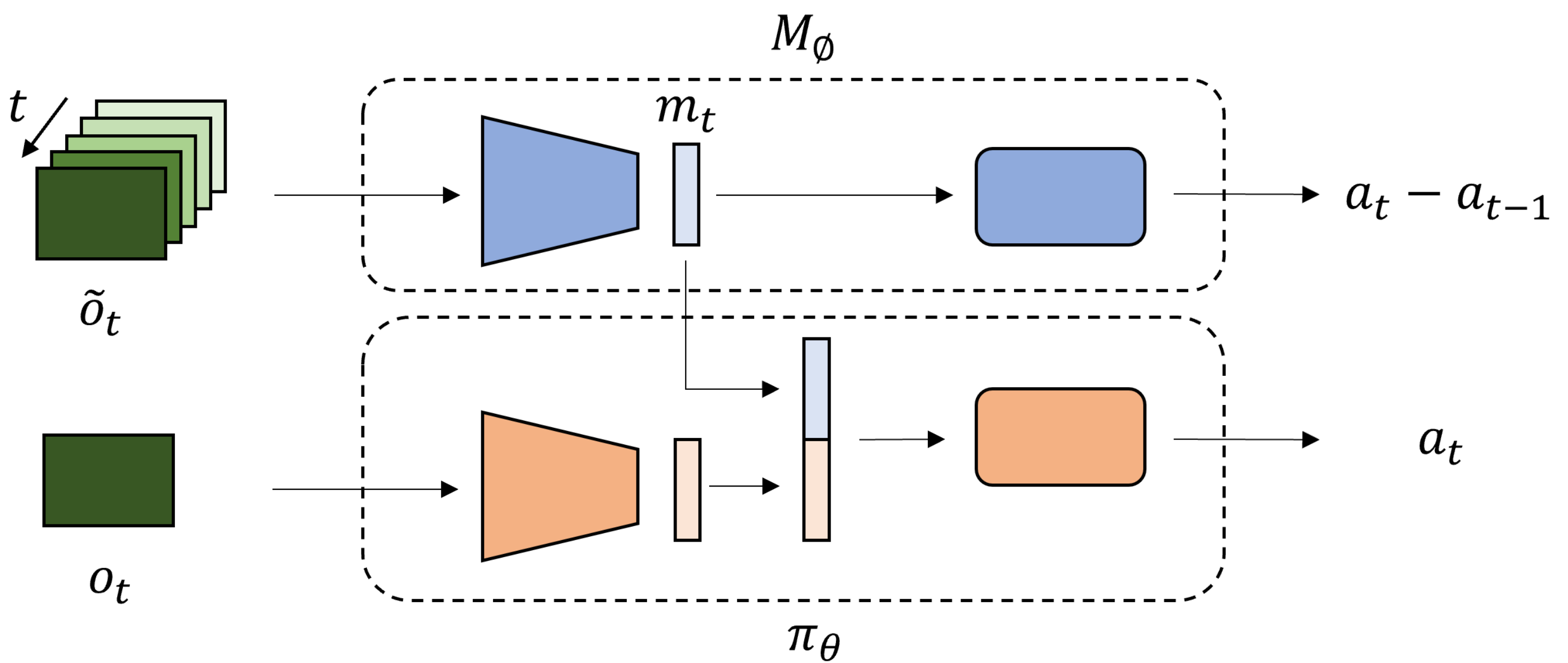}
    \caption{\textbf{Our framework}: We propose to solve the problem with one memory extraction stream (upper stream $M_\phi$) and a policy stream (lower stream $\pi_\theta$). The memory extraction stream extracts copycat-free historical features, and the policy stream fuse the current observation and the history to output final decisions. Furthermore, blocks with the same color indicate they are updated in the same optimization step.}
    \label{fig:framework}
\end{figure}

\subsection{Model Architecture}
\label{sec:model architecture}

Since the source of copycat problem is the nuisance information about $a_{t-1}$ implied in the temporal information in the observation history $\tilde{o}_{t}$, the principle of designing the imitation model architecture should be to extract as much information as possible from the observations to predict the action $a_{t}$ while ensuring that most information about $a_{t-1}$ is removed. Only the observation history $\tilde{o}_t$ contains information about $a_{t-1}$, and the current observation $o_t$ does not, so we split the neural network into two pathways: the $\tilde{o}_t$ pathway and the $o_t$ pathway. We propose a \textbf{memory extraction module} to extract history-related features $m_t$ from the $\tilde{o}_t$ input, such that it contains little information about $a_{t-1}$. Then we concatenate the feature $m_t$ and the current observation $o_t$ to predict the final action. We name the second network \textbf{policy network} since it fuses the history information and the current information to make the decision. Fig.~\ref{fig:framework} shows our framework, and we describe each of the components as follows.

\subsubsection{Memory Extraction Module}

The memory extraction module $M_{\phi}$ aims to get the additional copycat-free feature from $\tilde{o}_{t}$ and compensate for the missing information due to the partial observation when predicting the action $a_{t}$. We extract the intermediate embedding of $M_{\phi}$, named memory feature $m_{t}$, to represent the historical information. To make $m_{t}$ free from the copycat problem, we design a specific prediction objective for the memory extraction module $M_{\phi}$.
As discussed in Sec.~\ref{sec:copycat}, the copycat problem is mainly caused by the excessive information about $a_{t-1}$ in $\tilde{o}_{t}$. Motivated by this, $m_{t}$ is expected to contain the essential information for predicting the action $a_{t}$ while avoiding the copycat problem by retaining as little information about $a_{t-1}$ as possible. I.e. $M_{\phi}$ should be trained to maximize the mutual information between $m_{t}$ and $a_{t}$ while minimizing that between $m_{t}$ and $a_{t-1}$. This goal can be formalized as a maximizing conditional mutual information objective:
\begin{equation}
\label{eq:mutual-information}
    \phi^{*} = \arg\max_{\phi}I_{\phi}(m_{t};a_{t}|a_{t-1}).
\end{equation}

This conditional mutual information maximization objective allows the learned representation $m_t$ to contain as much unique information about $a_t$ as possible. Furthermore, since $a_{t-1}$ is in the condition, there is no incentive for $m_t$ to contain any information about $a_{t-1}$. Due to the limited model capacity and training data, more unique information about $a_t$ in $m_t$ results in the suppression of information about $a_{t-1}$. We remark that though this objective cannot guarantee to remove all information about $a_{t-1}$ from $m_t$, the results in Section~\ref{sec:analysis} indicate that it is sufficient to resolve most of the copycat problems in the visual tasks.

In practice, Eq.~\eqref{eq:mutual-information} is challenging to implement and unstable to optimize. Therefore, we propose to optimize over a lower bound on the conditional mutual information to learn the representation $m_t$ better.

\noindent\textbf{Theorem 1 (lower bound of Eq.~\eqref{eq:mutual-information}).} \textit{Let $r_t:=a_t - a_{t-1}$ be the action residual, then we have $I_{\phi}(m_{t},a_{t}|a_{t-1}) \geq \mathbf{H}(r_{t}|a_{t-1}) - \mathbf{H}_{\phi}(r_{t}|m_{t})$, where $\mathbf{H}$ represents the Shannon entropy.}

We can maximize this lower bound to approximate Eq.~\eqref{eq:mutual-information}. Because the first term is not correlated with the parameter $\phi$, the objective to optimize $M_{\phi}$ can be rewrited as $\phi^{*} = \arg\min_{\phi}H_{\phi}(r_{t}|m_{t})$. Furthermore, minimizing the conditional mutual information is equivalent to the maximum likelihood estimation, so we propose an \textbf{action residual prediction} objective for the memory extraction module $M_{\phi}$:
\begin{equation}
\label{eq:residual-predict}
    \phi = \arg\max_{\phi}\mathbb{E}_{\mathcal{D}}[ \log P(a_{t}-a_{t-1}|\tilde{o}_{t};\phi)].
\end{equation}

Intuitively, this approximated objective function aims to predict the change of actions. Since the expert actions in a decision process usually transit smoothly, predicting the action change approximately predicts the critical differences in the environments that need expert action change. 

Trained with Eq.~\eqref{eq:residual-predict}, the memory extraction module $M_{\phi}$ optimizes over the objective in Eq.~\eqref{eq:mutual-information} as well, i.e. extracting the sufficient information about $a_{t}$ and removing the shortcut information about $a_{t-1}$. Therefore, the memory feature $m_{t}$ provides additional information from history and cuts off the shortcut path $a_{t-1}\rightarrow a_{t}$ by removing most information about $a_{t-1}$. With the proposed architecture, the policy model $\pi_{\theta}$ will take the advantage of both the visual clues in $o_{t}$ and the complementary information provided by $m_{t}$ to learn the correct way for predicting $a_{t}$.

\subsubsection{Policy Module}

The policy module $\pi_{\theta}$ takes the current observation $o_{t}$ and the memory feature $m_{t}$ as input and predicts $a_{t}$, where $m_{t}$ is extracted from $M_{\phi}(\tilde{o}_{t})$ and contains additional and copycat-free historical information about predicting $a_{t}$. 
In practice, we process the observation with a convolutional neural network and fuse the mid-layer representation with the memory extraction module by concatenation. 
The objective of the policy model is:
\begin{equation}
\label{eq:pi-objective}
    \theta^{*} = \arg\max_{\theta} \mathbb{E}_{\mathcal{D}}[\log P(a_{t}|o_{t},m_{t};\theta)].
\end{equation}

\subsection{Implementation details}
\label{sec:implementation details}
In our implementation, we feed $o_{t}$ and $\tilde{o}_{t}$ into $\pi_{\theta}$ and $M_{\phi}$ respectively. We take the intermediate feature of $M_{\phi}$ as $m_{t}$ and then fuse it into $\pi_{\theta}$. Specifically, we directly concatenate $m_{t}$ with the intermediate feature of $\pi_{\theta}$ through a stop-gradient layer and feed them into the following layers of $\pi_{\theta}$. The respective targets of $\pi_{\theta}$ and $M_{\phi}$ are $a_{t}$ and $a_{t} - a_{t-1}$. $\pi_{\theta}$ and $M_{\phi}$ are trained jointly with the objectives Eq.~\eqref{eq:pi-objective} and Eq.~\eqref{eq:residual-predict}. We will introduce experiment setup and other implementation details in Sec.~\ref{sec:experiment-setup} and Appendix respectively.

\section{Experiments}

In this section, we aim to answer the following questions. 1) How well does our method performs? Does it outperform the previously proposed methods? 2) We would like to understand the role of the memory extraction module and provide empirical evidence that it indeed excludes information about $a_{t-1}$, thus resolving the copycat problem. 3) We would like to understand in what circumstances the proposed model is better qualitatively.

To answer these questions and evaluate our method, we conduct experiments in the CARLA~\cite{Dosovitskiy17} autonomous driving simulator and three standard OpenAI Gym MuJoCo \cite{todorov2012mujoco} continuous control environments: Hopper, HalfCheetah and Walker2D.

\subsection{Experiment Setup}
\label{sec:experiment-setup}

\textbf{CARLA\cite{Dosovitskiy17}:~~}CARLA is a photo-realistic autonomous driving simulator that features realistic driving scenarios. CARLA is also the most challenging and widely used environment to benchmark how well the copycat problem is resolved in the literature~\cite{wen2021keyframefocused}. We mostly follow the experimental setting from \cite{codevilla2019exploring}, but withhold the ego-agent velocity from the observation for all methods following \cite{wen2021keyframefocused} to ensure the environment is partially observed. The expert demonstration dataset is collected by a scripted expert. Under the hood, the expert knows the scene layout, including information such as the location and speed of other vehicles and pedestrians.

Each expert driving trajectory can be described by a list of $(o_t, c_t, a_t)$ tuples, where $t$ indexes the timestep. Here $o_t$ is the observation in the form of an RGB image. In our setup, the image is resized to $200\times 88$. $c_t$ is the driving command (one of the following commands: follow the lane, turn left, turn right, and go straight). The driving command is necessary because one has to command the autonomous driving agent about where to go. Finally, $a_t\in[-1,1]^2$ is a two-dimensional vector for controlling the steer and acceleration. Here the positive acceleration means throttle, and negative means brake. We use the CARLA100~\cite{codevilla2019exploring} dataset that consists of 100 hours of driving data. 
Table~\ref{tab:CARLA100} shows actions are smooth most of the time, so we can effectively verify whether the method can deal with the copycat problem. The BC agent aims to learn a function $f(\tilde{o}_t, c_t) \rightarrow a_t$. We follow previous setups~\cite{wen2021keyframefocused} and use $H=6$ for BCOH.

\begin{table}[htbp]
    \setlength{\abovecaptionskip}{-0.5cm}
    \caption{The distribution of $||a_t-a_{t-1}||^{2}_{2}$ in CARLA100}
    \label{tab:CARLA100}
    \centering
    \begin{tabular}{*{4}{c}}
        \hline
        $[0, 10^{-3})$  & $[10^{-3}, 10^{-2})$ & $[10^{-2}, 10^{-1})$ & $\geq 10^{-1}$\\
        \hline
        $68.8\%$ & $7.8\%$ & $14.3\%$ & $9.0\%$ \\
        \hline
    \end{tabular}
\end{table}

\noindent \textbf{CARLA \emph{NoCrash} benchmark \cite{codevilla2019exploring}:~~}
\emph{NoCrash} benchmark~\cite{codevilla2019exploring} is a series of navigation tasks consisting of 25 routes over 4 kinds of weather. Finishing a route means following the given route within the time constraint without collisions. There are three traffic conditions: \emph{Empty}, \emph{Regular}, and \emph{Dense} from easy to difficult. We focus on the \emph{Dense} and \emph{Regular} to ensure the agent learns the information about pedestrians or vehicles on the road. In addition to the same weathers and town as the training data, we evaluate the performance in the new town and the new weather. For each experiment, we count the number of successful episode denoted as \emph{\#SUCCESS} to be the main performance metric. We will describe model detail and other important metrics in Appendix.

\noindent \textbf{MuJoCo-Image(Hopper,Halfcheetah,Walker2D) 
\cite{todorov2012mujoco}:~~}
Following the same setting in \cite{wen2021keyframefocused}, we set the observation $o_t$ at time step $t$ to be a $128\times 128$ RGB image of the rendering environment and $H=1$ for BCOH. These control tasks have various state and action spaces, transition dynamics, and reward functions. Expert datasets are collected by a TRPO agent \cite{schulman2015trust} with access to fully observed states. Each dataset is a list of $(o_t,a_t)$ tuples, where $t$ denotes the timestep. We collected 1k samples for HalfCheetah, and 20k for Hopper and Walker2D. The BC agent is required to learn a function $f(\tilde{o}_t)\to a_t$. The performance metric is the average reward from the environment. We will report the details about the model architectures in Appendix.

\subsection{Previous Methods}

We compare our method to previous methods that are proposed to solve the copycat problem. We introduce those methods as follows.

\noindent\textbf{BCSO \& BCOH:~~} For CARLA, We use the model proposed in \cite{codevilla2019exploring} as the baseline model, and it was proposed along with CARLA \emph{Nocrash} benchmark. For MuJoCo, we follow the model in \cite{wen2021keyframefocused}.

\noindent\textbf{DAGGER \cite{Ross2011}:~~} \textbf{DAGGER} is a widely used online method in BC to deal with the causal confusion by querying experts to label new trajectories generated by the agent and then updating it. We note that this is considered an oracle method since DAGGER requires extra online interactions and expert labels. Therefore, we set the number of online queries to 150k for CARLA and 1k for MuJoCo.

\noindent\textbf{Historical-Dropout (HD) \cite{bansal2018chauffeurnet}:~~} \textbf{HD} randomly dropouts historical frames to cope with the copycat problem. In practice, we add a dropout layer with a probability of 0.5 on past observations  $o_{t-1}, o_{t-2},\cdots,o_{t-H}$.

\noindent\textbf{Fighting Copycat Agent (FCA) \cite{chuan2020fighting}:~~} \textbf{FCA} is an adversarial training approach to remove the information about the previous action in the observation histories. It encodes the observation histories to a latent vector, asks it to predict the current action with supervised loss, and asks it not to predict the previous action using adversarial loss. 

\noindent\textbf{Keyframe-Focused (Keyframe) \cite{wen2021keyframefocused} :~~} \textbf{Keyframe} re-weights the samples according to a weighting defined by how fast action changes. The places where action changes unexpectedly are considered as important timesteps.

Note that there are some other methods~\cite{chen2019lbc,ohn2020Learning,zhang2021end2end,chen2021learning} that achieve better performance on the CARLA benchmark.
However, our goal is to resolve the copycat problem, instead of achieving the best results on CARLA. Following prior works~\cite{wen2021keyframefocused}, we take \cite{codevilla2019exploring} as the base algorithm. We expect our techniques will also benefits more complicated imitation methods.

\subsection{Results}

\label{sec:results}

We report the mean and standard deviation of the performance metric for all entries and analyze the experiment results in this section.

\begin{table}[h]
     \setlength{\abovecaptionskip}{-0.5cm}
    \caption{The \#\textit{SUCCESS} on CARLA \emph{Nocrash} benchmark. For each method, we train 3 policies from different initial seeds.}
    \label{tab:CARLA-results}
    \centering
    \begin{tabular}{l| *{2}{c} | *{2}{c} | *{2}{c}}
         \hline
         Environment & \multicolumn{2}{c}{Train Town \& Weather} \vline & \multicolumn{2}{c}{New Weather} \vline& \multicolumn{2}{c}{New Town} \\
         \hline
         Traffic & \emph{Regular} & \emph{Dense} & \emph{Regular} & \emph{Dense} & \emph{Regular} & \emph{Dense} \\
         \hline
         BCSO & $37.6\pm6.1$ & $13.1\pm1.8$ & $18.3\pm1.7$ & $5.7\pm5.7$ & $10.3\pm2.9$ & $2.0\pm0.8$ \\
         BCOH & $67.1\pm10.8$ & $34.1\pm7.5$ & $26.0\pm6.5$ & $6.0\pm2.9$ & $27.0\pm5.0$ & $15.3\pm2.9$  \\
         OURS & \bm{$78.1\pm0.9$} & \bm{$52.0\pm2.3$} & \bm{$39.3\pm5.8$} & \bm{$19.0\pm2.7$} & \bm{$40.7\pm7.0$} & \bm{$25.7\pm4.2$}  \\
         \hline
         DAGGER & $69.7\pm8.4$ & $42.7\pm5.7$ & $24.7\pm4.8$ & $14.7\pm4.2$ & $34.0\pm4.3$ & $12.0\pm4.5$  \\
         HD & $70.1 \pm 4.0$ & $35.6\pm3.5$ & $28.0\pm6.5$ & $16.7\pm6.2$ & $32.0\pm4.3$ & $11.3\pm2.5$ \\
         FCA & $58.0 \pm 8.0$ &$31.2\pm5.2$ & $20.0\pm11.5$ & $8.7\pm4.0$ & $21.3\pm3.1$ & $8.3\pm2.9$ \\
         Keyframe & $74.4 \pm 7.3$ & $41.9\pm6.2$ & $34.0\pm5.4$ & $13.7\pm5.4$ & $33.3\pm7.3$ & $16.3\pm3.9$  \\
         \hline
    \end{tabular}
\end{table}

\begin{figure}[ht]
    \setlength{\belowcaptionskip}{-0.5cm}
    \centering
    \includegraphics[width=0.8\textwidth]{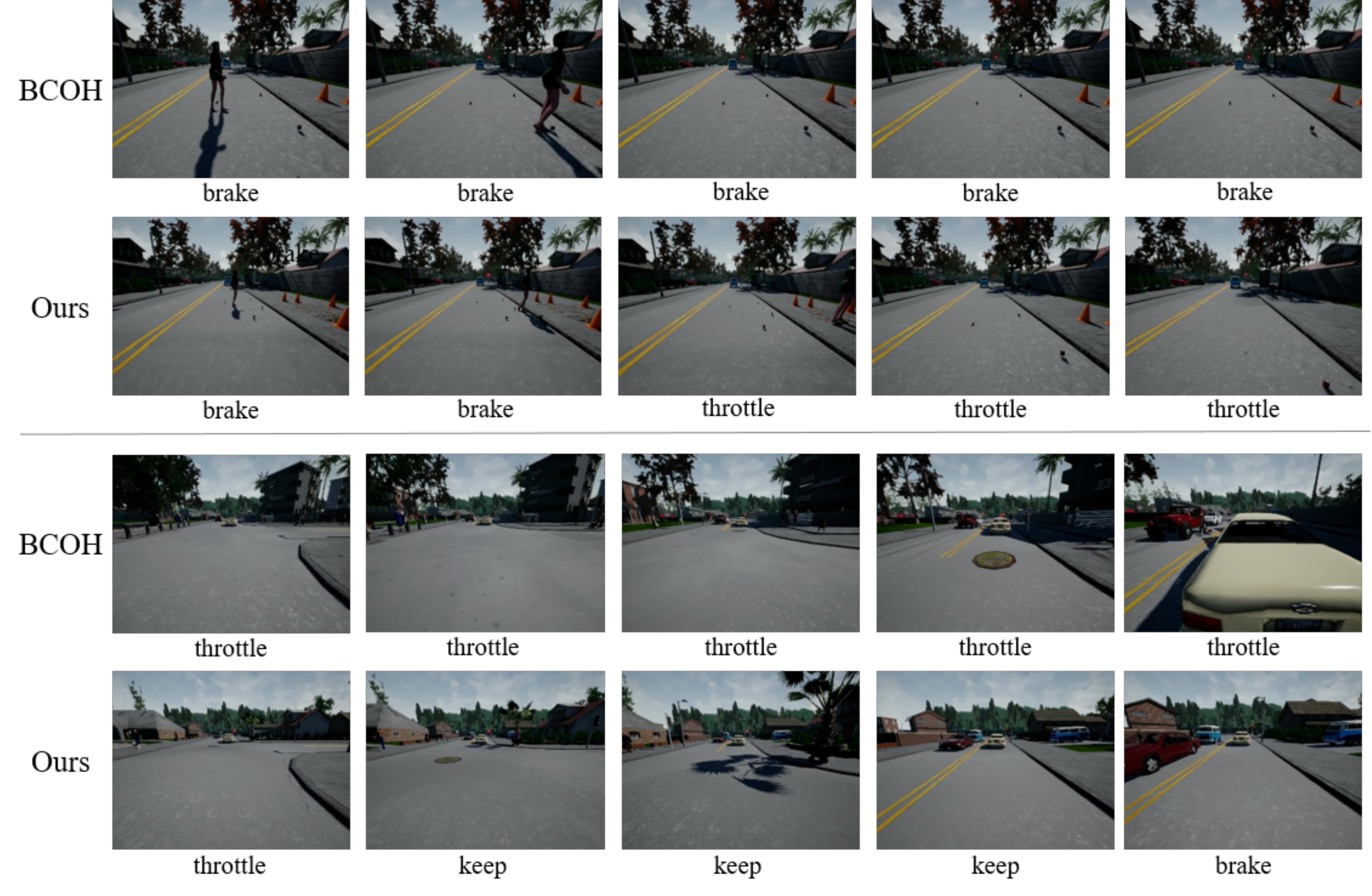}
    \caption{Visualizations of BCOH and our method in CARLA: Each row is a video sequence. The first two rows show a case where BCOH is stopped for a pedestrian and never restarts, while our method handles that successfully. The last two rows show a case where BCOH passes an intersection and then fails to stop behind a vehicle while our method executes the proper behavior. The ``throttle/keep/brake" indicates the acceleration of the agent.}
    \label{fig:CARLA copycat}
\end{figure}

CARLA \emph{Nocrash} benchmark results are shown in the Table~\ref{tab:CARLA-results}. We conclude that our method achieves the best performance compared to all previous approaches, including the oracle approach DAGGER. Besides our superior performance, we would also like to describe the results qualitatively. BCSO in CARLA severely fails because the agent can not observe the motion of the surrounding agent, as well as that of itself. Thus, BCOH is necessary in this case. We observe severe copycat problems with BCOH. A lot of failure cases for BCOH are when it is stopped for some obstacle, and it cannot restart again due to the copycat issue. This leads to a high timeout failure rate. Fig.~\ref{fig:CARLA copycat} shows the copycat problem of BCOH and these are the main reasons for the poor performance. The oracle method DAGGER is better than baseline in the \emph{Dense} traffic case but similar to baseline in \emph{Regular} traffic. We hypothesize that the \emph{Dense} traffic evaluation has a significant distributional shift compared to the training data. HD performs similarly to BCOH, indicating it can not solve the copycat problem in this challenging case. FCA fails to outperform the BCOH baseline because it can not handle the large complex input space with the adversarial training mechanism which was also mentioned in \cite{wen2021keyframefocused}. Keyframe significantly outperforms the BCOH baseline but is still worse than our method.

MuJoCo-Image results are shown in Table~\ref{tab:MuJoCo-results}. BCOH yields higher rewards than BCSO due to access to motion information, and the performance can be further improved by addressing the copycat problem. DAGGER performs well on MuJoCo and gets the best results in Hopper. HD fails in these settings and has a similar performance to BCSO, indicating it cannot resolve the copycat problem. FCA yields a lower reward than BCOH because it is difficult to handle high-dimensional input with adversarial learning. For the other offline method Keyframe, though it significantly improves the performance, ours outperforms it in Walker2D and HalfCheetah and achieves a comparable performance in Hopper. Results on more MuJoCo environments are shown in Appendix.

\begin{table}[h]
    \caption{The average reward in MuJoCo-Image. For each method and task, we train 5 policies from different initial seeds.}
    \label{tab:MuJoCo-results}
    \centering
    \resizebox{\textwidth}{!}{
    \begin{tabular}{l| *{3}{c} | *{4}{c}}
      \hline
      Method & BCSO & BCOH & OURS & DAGGER & HD & FCA & Keyframe \\
      \hline
      Hopper & $601\pm168$ & $740\pm35$ & $894\pm38$ & \bm{$1034\pm45$} & $617\pm111$ & $735\pm106$ & $951\pm117$ \\
      Walker2D & $481\pm40$ & $614\pm107$ & \bm{$800\pm58$} & $699\pm111$ & $594\pm61$ & $534\pm99$ & $769\pm91$ \\
      HalfCheetah & $4\pm5$ & $615\pm41$ & \bm{$914\pm115$} & $822\pm186$ & $96\pm40$ & $270\pm168$ & $819\pm96$ \\
      \hline
    \end{tabular}
    }
\end{table}

\subsection{Ablation Studies}
\label{sec:Ablation Studies}

To evaluate the effect of each module and technique we propose in Sec.~\ref{sec:model architecture}, we conduct ablation study experiments in CARLA \emph{Nocrash} \emph{Dense}. The results are shown in Table \ref{table:ablation}, and we provide detailed analysis for each ablation, respectively.

\begin{table}[h]
\setlength{\abovecaptionskip}{-0.5cm}
\caption{Ablation results in CARLA \emph{Nocrash} \emph{Dense}}
\label{table:ablation}
\centering
\begin{tabular}{l|c}
\hline 
    \textbf{Architecture} & \emph{\#Success}($\uparrow$)\\
    \hline
    Ours & \bm{$52.0\pm2.3$}\\
    \hline
    Memory only: residual controller & $0.0\pm0.0$\\
    Memory only: learned controller & $0.0\pm0.0$\\
    \hline
    Memory module objective: $a_t$ & $41.3\pm1.9$\\
    Memory module objective: $a_{t-1}$ & $47.0\pm5.1$ \\
    \hline
    Without stop-gradient layer & $45.0\pm5.3$\\
    \hline
\end{tabular}
\end{table}

\noindent \textbf{The memory extracted module only:~~}
We want to show both parts of our framework are necessary. From the BCSO result, we know the policy module alone achieves inferior performance. To evaluate the memory extraction module alone, we set up two experiments that only uses the well-trained memory extracted module to derive the action: one uses the residual directly (\textbf{Memory only: residual controller}), i.e., it uses the output of the memory extracted module plus the previous action to get the current action $a_t=a_{t-1}+r_t$; the other uses the extracted feature (\textbf{Memory only: learned controller}) to control the vehicle with the same controller as the baseline model, i.e., the controller is trained by using $m_t$ from memory extracted module to fit $a_t$. However, the imitators in both setups fail in every episode. It indicates the memory extracted feature $m_t$ alone contains insufficient information for the task. The residual prediction not only removes the information about $a_{t-1}$ but also erases some necessary information about the current scene. Therefore, the policy module completes information related to control by adding the current observation $o_t$. This illustrates that the policy stream is an indispensable part of our framework.

\noindent \textbf{Alternative memory module objectives:~~}
Moreover, we would like to investigate the performance of alternative memory module designs. Our design is based on a lower bound of the conditional mutual information objective. Here we train the model from scratch by replacing the memory module prediction objective to $a_t$ and $a_{t-1}$ respectively. As shown in \textbf{memory module objective: $a_t$} and \textbf{memory module objective: $a_{t-1}$}, both of the objective changes lead to deterioration of the performances. This indicates that other features are less informative or more likely to suffer from the copycat problem.

\noindent \textbf{Without stop-gradient layer:~~}
The stop-gradient is applied to the memory feature $m_t$ to ensure the policy module doesn't directly learn shortcut information from the observation histories. After removing it, the result shows the performance degradation, indicating that the imitation learner might learn some copycat information from the observation histories. Though the action residual prediction objective removes some shortcut information about $a_{t-1}$ and improves the performance compared to BCOH, the stop-gradient is indispensable for the policy module to cut off the nuisance information flow from observation histories.

\subsection{Analysis}
\label{sec:analysis}

The performance of our framework is clearly above that of alternative methods on the CARLA \emph{Nocrash} benchmark. In this subsection, we prove that the training objective proposed in Eq.~\eqref{eq:residual-predict} can successfully resolve the copycat problem. We design two analytical experiments to evaluate our method qualitatively. The first analysis directly proves our method has fewer copycat behaviors by an intervention experiment. The second one validates our training objective indeed reduces the mutual information between $a_{t-1}$ and $m_t$. We choose BCOH and Keyframe as the baselines for these two analytical experiments.

\begin{table}[h]
\setlength{\abovecaptionskip}{-0.5cm}
\caption{The analysis results on the validation set}
\label{tab:analysis-results}
\centering
\begin{tabular}{l|*{3}{c}}
    \hline 
    Method & BCOH & Keyframe & OURS \\
    \hline
    Change(\%)($\downarrow$) & 40.88 & 22.96 & \textbf{16.18}\\
    \hline
    MSE loss $\times 10^{-2}$($\uparrow$) & 4.83 & 5.04 & \textbf{7.50} \\
    \hline
\end{tabular}
\end{table}

\subsubsection{Intervention on Observation Histories} 
To verify whether our model learns the causal effect of the observation histories $\tilde{o}_t$, we propose an intervention analysis. We replace the original $\tilde{o}_t$ with a counterfactual history $do(\tilde{o}_t)=[o_t,o_t,\cdots,o_t]$, i.e., all histories are the current observation $o_t$.For a copycat agent, then with the counterfactual observation histories $do(\tilde{o}_t)$ as the input, it will have an illusion that the vehicle is in the stationary state and will merely repeat the previous motion of staying still. A causally correct agent should perform according to the content of the scene. We test our model and the baseline models under the cases where the vehicles have positive speed and the latest action is non-stop (denoted by $\pi(\cdot) > 0$ with a slight abuse of notation). A causally correct agent should drive forward in this case, i.e. $\pi(do(\tilde{o}_t))>0$. We count the percentage where the model stopped on the intervened observation, i.e., $\frac{N(v>0, \pi(\tilde{o}_t)>0, \pi(do(\tilde{o}_t))=0)}{N(v>0, \pi(\tilde{o}_t)>0)}$  where $N(.)$ denotes the counting function. The behavior change percentage can be regarded as the error rate of this experiment.

The 2nd row of Tab.~\ref{tab:analysis-results} shows the intervention results. As shown in the table, $40.88\%$ of samples change in BCOH, even there is no need for the agent to stop in these scenes (e.g., no vehicles or pedestrians ahead; the traffic signal is not red). It illustrates that BCOH makes decisions by merely repeating the previous action but ignoring the perception of the current scene in nearly half of the cases. The metric of Keyframe is relatively lower than that of BCOH, which indicates Keyframe does not rely just on the previous action and partially mitigates the copycat problem. Meanwhile, OURS is only $16.18\%$, showing that our approach significantly alleviates the copycat problem.

\subsubsection{Mutual Information} 
In Sec.~\ref{sec:method}, we set up a conditional mutual information objective, i.e. maximize $I(m_t; a_t | a_{t-1})$. And we use an approximation to this objective during the optimization process. In this analysis, we empirically evaluate how much information about $a_{t-1}$ is left in $m_t$. This would help to justify both the objective function as well as the approximation. More concretely, we regress $a_{t-1}$ from the representation with a 2-layer MLP. We calculate the validation loss of $a_{t-1}$ and use it as a proxy to the negative of the mutual information between the representation and $a_{t-1}$. The less the information, the less likely the method would suffer from the copycat issue. We evaluate our method's representation $m_t$ and corresponding layers of the BCOH and the Keyframe methods. 

The results in the 3rd row of Tab.~\ref{tab:analysis-results} show that the feature in BCOH is more predictive of the previous action. This conclusion matches the preliminary results that BCOH is more prone to copycat problems and learns the spurious temporal correlation. The MSE of Keyframe is slightly larger than that of BCOH, which shows that this method removes part of the information about the previous action and mitigates the copycat problem. Meanwhile, the MSE of OURS is the largest, which indicates that the extracted feature $m_t$ in our model contains the smallest amount of mutual information with $a_{t-1}$ and cuts off the shortcut of simply imitating the previous action from observation histories.

\section{Conclusion}
Behavioral cloning is a fundamental algorithm that helps the agent to imitate complex behaviors. However, it suffers from causal issues due to the temporal structure of the problem. We view the causal issue from an information flow perspective and propose a simple yet effective method to improve performance drastically. Our method outperforms previous solutions to the copycat problem. In the future, we would like to investigate casual issues in a broader range of algorithms, e.g., reinforcement learning and other sequential decision problems. 

\section{Acknowledgement}
We thank Jiaye Teng, Renhao Wang and Xiangyue Liu for insightful discussions and comments.
This work is supported by the Ministry of Science and Technology of the People's Republic of China, the 2030 Innovation Megaprojects ``Program on New Generation Artificial Intelligence" (Grant No. 2021AAA0150000), and a grant from the Guoqiang Institute, Tsinghua University.

\clearpage
%
%
\bibliographystyle{splncs04}
\bibliography{egbib}

\newpage
\appendix
\addcontentsline{toc}{section}{Appendices}
\section*{Appendices}

\section{Proof of Theorem 1}

$I_{\phi}(m_{t};a_{t}|a_{t-1})$ is the parametrized conditional mutual information between $m_t$ and $a_{t}$ on the condition of $a_{t-1}$. The first equality holds since $a_t = r_t + a_{t-1}$. Then, the second equality can be obtained by using definitions of mutual information to expand $I_{\phi}(m_{t};r_{t}|a_{t-1})$. Note that the conditional entropy $\mathbf{H}(r_{t}|a_{t-1})$ is not related to our optimizing variables $\phi$ since it doesn't contain $m_t$. Furthermore, according to the total probability formula, we can expand $\mathbf{H}_{\phi}(m_{t},r_{t}|a_{t-1})$ to eliminate $\mathbf{H}_{\phi}(m_{t}|a_{t-1})$ and derive the third equality. The final inequality holds since the conditions $\{m_t,a_{t-1}\}$ is a superset of the conditions $\{m_t\}$. 

\begin{equation*}
\begin{aligned}
    & I_{\phi}(m_{t};a_{t}|a_{t-1}) \\
    &= I_{\phi}(m_{t};r_{t}|a_{t-1}) \\
    &= \mathbf{H}_{\phi}(m_{t}|a_{t-1}) + \mathbf{H}(r_{t}|a_{t-1})- \mathbf{H}_{\phi}(m_{t},r_{t}|a_{t-1}) \\
    &= \mathbf{H}(r_{t}|a_{t-1}) - \mathbf{H}_{\phi}(r_{t}|m_{t},a_{t-1}) \\
    &\geq \mathbf{H}(r_{t}|a_{t-1}) - \mathbf{H}_{\phi}(r_{t}|m_{t}) \\
\end{aligned}
\end{equation*}

\section{Implementation Details of Experiments in CARLA}

\subsection{Architectural details \& Loss functions} 
We use the backbone of conditional imitation learning framework CILRS \cite{codevilla2019exploring} and set all the input speed $v_{in}$ to zero to create a POMDP \cite{wen2021keyframefocused}.

The input $o_t$ and $\hat{o}_t$ of all models is a three-dimensional tensor with the size of $30\times288\times80$. We stack the observed images ($3\times288\times80$ RGB images) along the first dimension in chronological order and set the total number of channels of all input tensors to 30 for fairness. $o_t$ contains only the current frame and $\hat{o}_t$ has a relatively long observation history. However, both $o_t$ and $\hat{o}_t$ have less than 10 images, so we set the remaining channels to all zeros. 

We use ImageNet-pretrained ResNet34 \cite{he2015deep} as the perception backbone for all methods to obtain latent representation. To accommodate 30-channel input, we repeat the first-layer convolution kernel 10 times in the first dimension and normalize the pretrained weight to $1/10$ of the original.

\begin{figure}[h]
    \centering
    \includegraphics[width=0.6\textwidth]{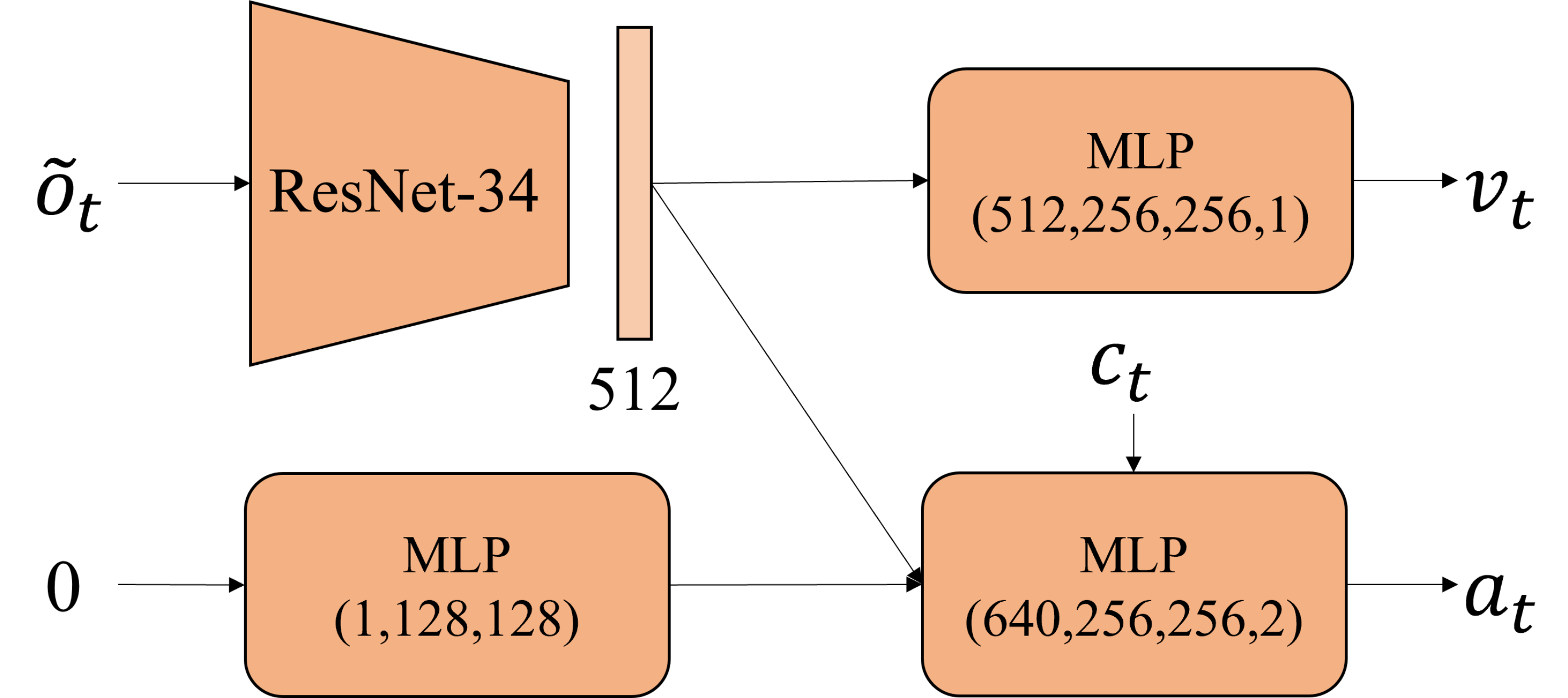}
    \caption{CILRS architecture: The model is used as the BCOH.}
    \label{fig:BCOH-detail}
\end{figure}

The details of BCOH are shown in Fig. \ref{fig:BCOH-detail}. Resnet34 casts the input $\hat{o}_t$ into a 512-dimensional compact representation. This representation is fed into a 3-layer MLP to obtain the estimated ego-velocity $v_t$ (a scalar). Besides, the representation is concatenated with the output of 2-layer MLP with all-zero input. Then the concatenated feature is fed into a 1-layer MLP which reduces its dimension to 512. This fusion 512-dimensional vector is then fed into the corresponding 3-layer MLP conditioned on the current time-step command $c_t$, which finally outputs the current action $a_t$ (a 2-dimensional vector). BCOH uses the speed regularization \cite{codevilla2019exploring} to address the causal confusions to some extent. Thus, the loss function for BCOH is defined as follows,
\begin{align}
    L_{\text{BCOH}} = \alpha L(a_t, a_t^{gt}) + (1-\alpha) L(v_t, v_t^{gt}),
    \label{eq: loss for BCOH}
\end{align}
where $a_t^{gt}$ and $v_t^{gt}$ are the ground truths of the current action $a_t$ and the speed $v_t$ respectively, $\alpha$ denotes the weighting to the loss of $a_t$, and $L$ is an L1 loss function.

\begin{figure}[h]
    \centering
    \includegraphics[width=0.6\textwidth]{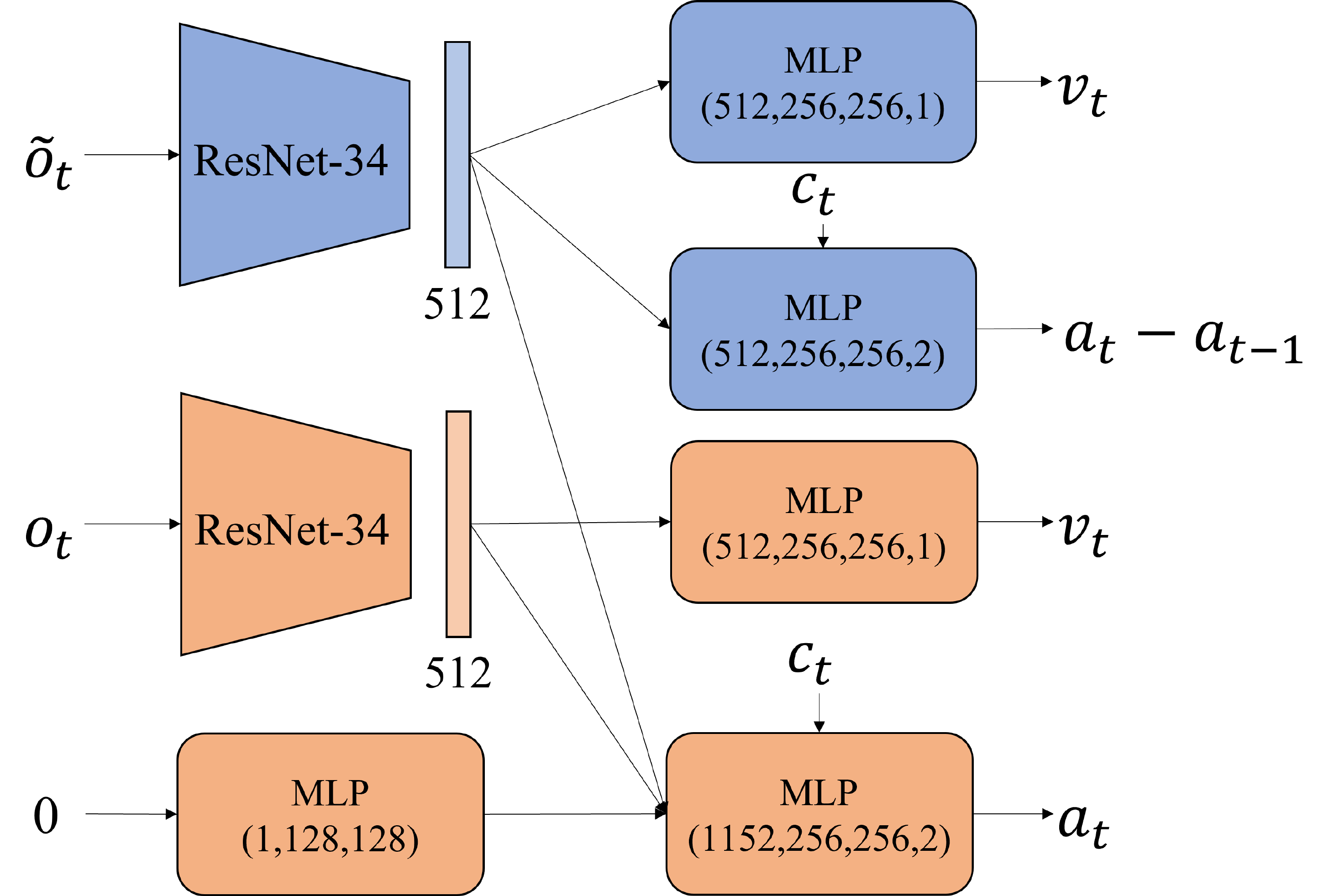}
    \caption{Our architecture: blue blocks are the memory extraction module; orange blocks are the policy module. Each module is a variant of the CIRLS architecture.}
    \label{fig:OURS-detail}
\end{figure}

The details of our model are shown in Fig.\ref{fig:OURS-detail}. The memory module and the policy module in our model share a similar architecture with BCOH's described above. However, the memory module removes the MLP for all-zero input $v_{in}$, and the input for the policy module is $o_t$. The basic training objectives of policy module $\pi_\theta$ and memory extraction module $M_\phi$ are $a_t$ and $a_t-a_{t-1}$ respectively. Similar to BCOH, each module of our model uses speed regularization. Therefore, the loss functions we designed for each module are:
\begin{equation}
    \begin{aligned}
        L_{M_\phi} &= \alpha L(a_t-a_{t-1}, a_t^{gt}-a_{t-1}^{gt}) + (1-\alpha) L(v_t, v_t^{gt}),\\
        L_{\pi_\theta} &= \alpha L(a_t, a_t^{gt}) + (1-\alpha) L(v_t, v_t^{gt}),\\
        L_{\text{overall}} &= L_{M_\phi} + L_{\pi_\theta}
    \end{aligned}
    \label{eq: loss for ours}
\end{equation}
where $a_{t-1}^{gt}$ is the ground truth of the previous action $a_{t-1}$ and other symbols are the same with those in Eq.\eqref{eq: loss for BCOH}.

\subsection{Architectural details of baselines in Ablation Studies}

\begin{figure}[h]
    \centering
    \includegraphics[width=0.6\textwidth]{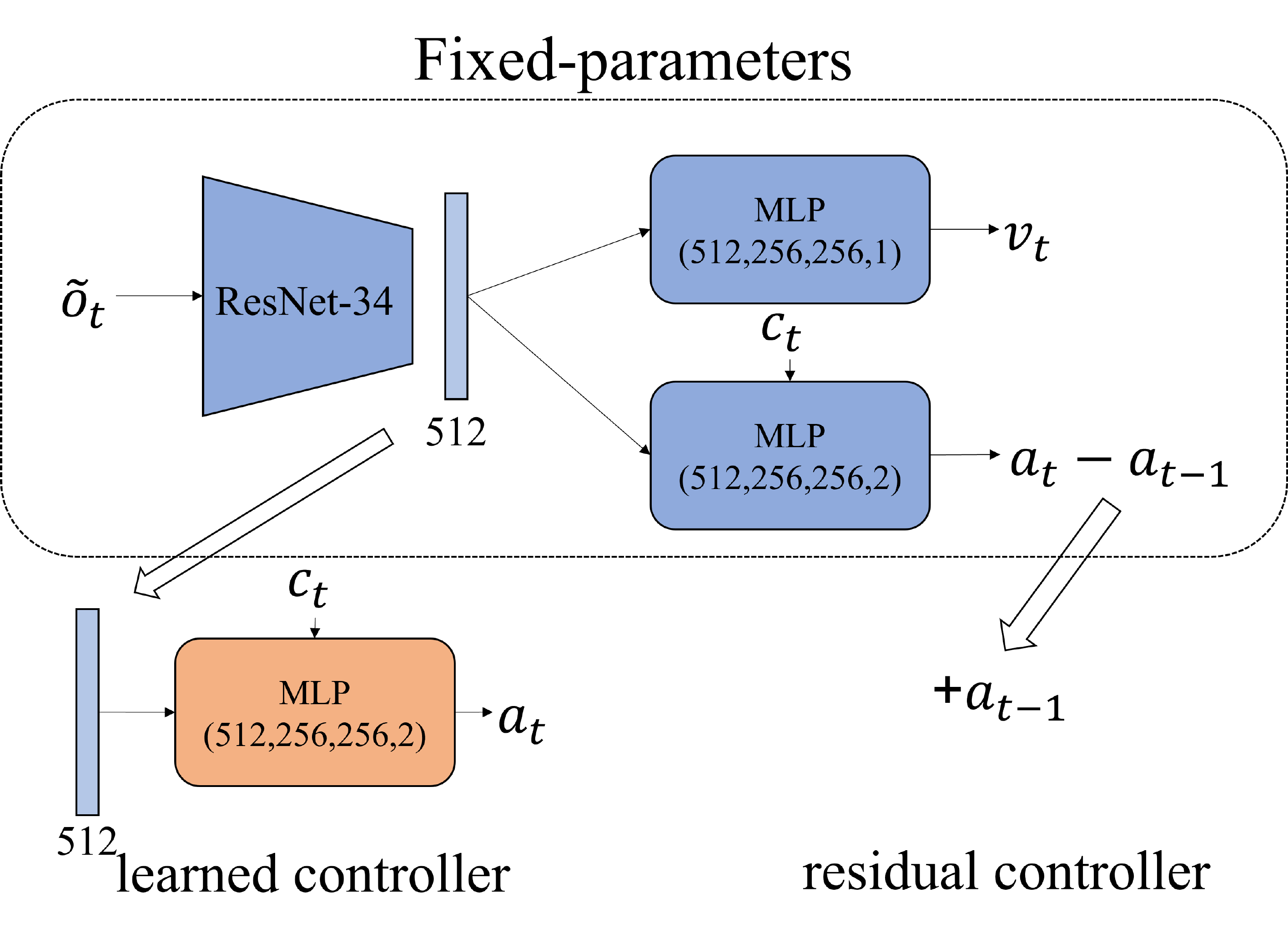}
    \caption{Memory only details}
    \label{fig:memory-only}
\end{figure}

Fig.~\ref{fig:memory-only} shows the details of \textbf{Memory only}. We fixed the parameters of a well-trained memory extracted module and try to use this module's output or intermediate feature to predict the current action $a_t$. The \textbf{Memory only: residual controller} adds the predicted residue output directly into last-step action $a_{t-1}$ to obtain the prediction of $a_t$. The \textbf{Memory only: learned controller} uses the extracted feature (the output of ResNet-34) as the input to regress $a_{t-1}$ via a 3-layer MLP.

\begin{figure}[H]
    \centering
    \includegraphics[width=0.6\textwidth]{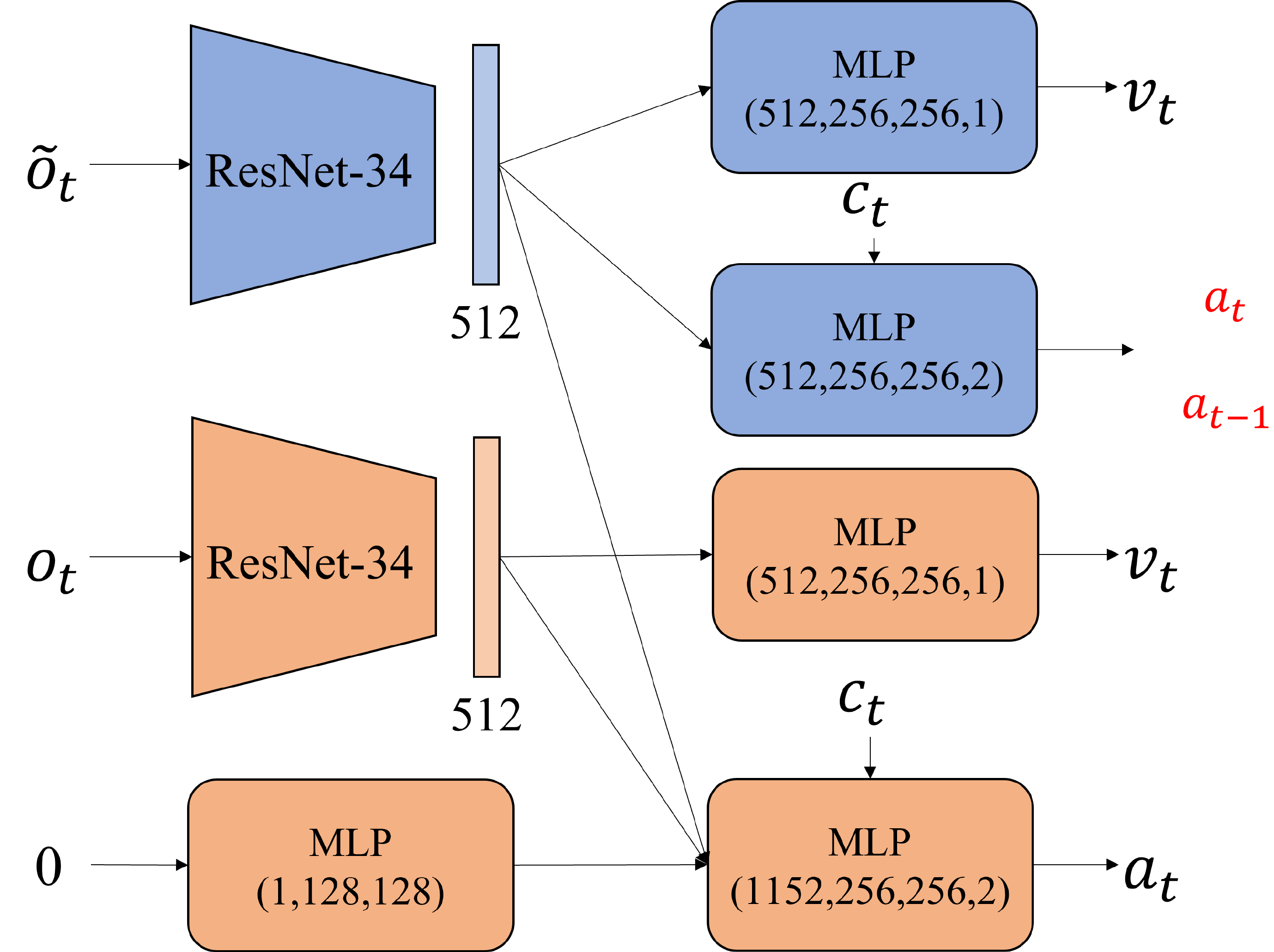}
    \caption{Memory module objective details}
    \label{fig:memory-objective}
\end{figure}

Fig.~\ref{fig:memory-objective} shows the details of \textbf{Memory module objective}. We train the model with different objectives ($a_t$ or $a_{t-1}$) for the memory extraction module, and the remaining setup is the same with the our proposed model.

\subsection{Other Details}
\label{sec: other details}
For all implemented methods, we apply the same hyper-parameters shown in Table~\ref{tab:hyper-params}, including total training iterations, batch size, $\alpha$, loss function, optimizer setup, and other configurations about the learning rate (LR) scheduling. 

\begin{table}[h]
    \caption{Hyper-parameters of experiments}
    \label{tab:hyper-params}
    \centering
    \begin{tabular}{l|c}
         Configuration & Value\\
         \hline
         Total training iterations & 100k \\
         Batch size & 160 \\
         $\alpha$ & 0.95 \\
         Loss function & $L_1$ \\
         \hline
         Optimizer & Adam \\
         Betas & (0.9, 0.999)\\
         Eps & 1e-08 \\
         Weight decay & 0 \\
         \hline
         Initial LR& 2e-4 \\
         LR decay threshold & 5000 \\
         LR decay rate & 0.1 \\
         LR lower bound & 1e-7 \\
         \hline
    \end{tabular}
\end{table}

\noindent \textbf{LR scheduling:~~} LR starts with an initial learning rate (initial LR) and decays when the best loss is unable to go down further for a preset number of iterations (LR decay threshold). Then, each decay learning rate is multiplied by a decay rate (LR decay rate) until it is lower than the set minimum learning rate (LR lower bound). As a result, the LR  adjusts adaptively and will not vanish in the whole training process.

\noindent \textbf{Data Augmentation:~~} We apply noise injection \cite{laskey2017dart} and multi-camera data augmentation \cite{bojarski2016end,giusti2015machine} on our training dataset to alleviate the distribution shift. Both of them are commonly used in the autonomous driving.

\noindent \textbf{Random seeds:~~}We retrained the proposed framework 3 times with different random initialization and test our agent on 25 routes for 4 kinds of weather with 3 different seeds. It makes sure we obtain a statistically significant better result.

\subsection{Failure mode in CARLA \emph{NoCrash}}
\label{appendix:failure mode}
\begin{table}[H]
    \caption{Failure mode on training conditions.}
    \label{tab:full-results}
    \centering
    \begin{tabular}{l| *{2}{c} | *{2}{c}}
         \hline
         Traffic & \multicolumn{2}{c}{\emph{Regular}} \vline & \multicolumn{2}{c}{\emph{Dense}} \\
         \hline
         Method  & \emph{\#COLLISION} & \emph{\#TIMEOUT}  & \emph{\#COLLISION}& \emph{\#TIMEOUT} \\
         \hline
         BCSO  & $53.0\pm7.9$ & \bm{$10.2\pm3.1$}  & $76.4\pm3.5$ & \bm{$11.1\pm2.9$}  \\
         BCOH  & $11.1\pm3.1$ & $21.9\pm12.7$  & $30.2\pm7.9$ & $36.1\pm14.5$  \\
         OURS  & \bm{$6.8\pm1.3$} & $15.2\pm0.2$  & \bm{$25.0\pm5.4$} & $23.3\pm7.6$  \\
         \hline
         DAGGER  & $14.8\pm2.9$ & $15.9\pm8.5$  & $35.0\pm3.6$ & $23.0\pm7.1$  \\
         HD  & $18.3 \pm 5.2$ & $12.2 \pm 4.4$  & $45.3\pm3.5$ &$20.3\pm5.6$  \\
         FCA  & $14.7 \pm 3.3$ & $27.3 \pm 8.8$  & $34.4\pm8.1$ & $35.3\pm9.6$  \\
         Keyframe  & $13.8 \pm 2.7$ & $11.9\pm5.8$  & $33.9\pm6.6$ &$24.8\pm7.9$  \\
         \hline
    \end{tabular}
\end{table}

\begin{table}[H]
    \caption{Failure mode on new weather}
    \label{tab:full-results-new-weather}
    \centering
    \begin{tabular}{l| *{2}{c} | *{2}{c}}
         \hline
         Traffic & \multicolumn{2}{c}{\emph{Regular}} \vline & \multicolumn{2}{c}{\emph{Dense}} \\
         \hline
         Method  & \emph{\#COLLISION} & \emph{\#TIMEOUT}  & \emph{\#COLLISION}& \emph{\#TIMEOUT} \\
         \hline
         BCSO  & $31.7\pm5.8$ & $8.7\pm3.1$  & $42.3\pm0.9$ & \bm{$6.3\pm1.2$}  \\
         BCOH  & $7.0\pm1.4$ & $16.0\pm6.4$  & $18.3\pm4.6$ & $16.3\pm6.8$  \\
         OURS  & \bm{$5.7\pm1.5$} & \bm{$3.7\pm4.7$}  & \bm{$18.0\pm2.6$} & $6.3\pm3.2$  \\
         \hline
         DAGGER  & $12.0\pm1.4$ & $10.7\pm1.7$  & $22.7\pm2.6$ & $13.3\pm7.1$  \\
         HD  & $11.0 \pm 2.8$ & $11.3 \pm 7.6$  & $21.0\pm3.6$ &$12.3\pm6.2$  \\
         FCA  & $9.0 \pm 2.2$ & $22.3 \pm 13.9$  & $18.7\pm9.6$ & $23.0\pm12.3$  \\
         Keyframe  & $7.3 \pm 1.2$ & $9.3\pm6.2$  & $22.7\pm2.9$ &$11.7\pm6.6$  \\
         \hline
    \end{tabular}
\end{table}

\begin{table}[H]
    \caption{Failure mode on on new town}
    \label{tab:full-results-new-town}
    \centering
    \begin{tabular}{l| *{2}{c} | *{2}{c}}
         \hline
         Traffic & \multicolumn{2}{c}{\emph{Regular}} \vline & \multicolumn{2}{c}{\emph{Dense}} \\
         \hline
         Method  & \emph{\#COLLISION} & \emph{\#TIMEOUT}  & \emph{\#COLLISION}& \emph{\#TIMEOUT} \\
         \hline
         BCSO  & $52.0\pm2.2$ & $30.3\pm0.9$  & $73.0\pm1.6$ & \bm{$22.3\pm1.7$}  \\
         BCOH  & $33.0\pm7.5$ & $42.0\pm13.6$  & \bm{$43.3\pm11.1$} & $52.0\pm13.4$  \\
         OURS  & \bm{$32.7\pm6.7$} & \bm{$28.0\pm4.6$}  & $50.3\pm4.7$ & $30.7\pm3.1$  \\
         \hline
         DAGGER  & $31.3\pm4.2$ & $36.0\pm7.5$  & $52.3\pm3.7$ & $36.7\pm6.6$  \\
         HD  & $30.7 \pm 4.2$ & $37.7 \pm 1.7$  & $55.3\pm6.3$ &$34.0\pm6.5$  \\
         FCA  & $31.3 \pm 9.1$ & $48.3 \pm 10.3$  & $49.0\pm8.6$ & $43.3\pm10.5$  \\
         Keyframe  & $34.3 \pm 1.2$ & $31.7\pm4.1$  & $48.3\pm3.9$ &$38.0\pm6.7$  \\
         \hline
    \end{tabular}
\end{table}

There are two kinds of failure modes in CARLA \emph{NoCrash}: collision and timeout. The collision means the driving agent falls the episode due to collision with other objects such as vehicles, pedestrians, and guardrails; The timeout means it exceeded the time limit of the episode. Failure mode results in CARLA \emph{NoCrash} are shown in Tab.~\ref{tab:full-results}, Tab.~\ref{tab:full-results-new-weather}, and Tab.~\ref{tab:full-results-new-town}. We note that our method is not always the lowest for the timeout failure rate, and that is because other methods might have a much higher collision rate. For example, BCSO is consistently the best in \emph{\#TIMEOUT} metric because most of its episodes end with collisions. Severe copycat problems with BCOH also lead to a high timeout failure rate.

\subsection{Other Experiments}
\noindent \textbf{Reactions to traffic lights~~}
Traffic lights are essential facilities for driving, and it decides whether the vehicle can pass the intersections safely. However, a traffic light occupies only a few pixels of the entire picture, and if it changes, it's hard for the imitation learner to concentrate on this slight but important change. Moreover, suppose the imitation learner suffers from copycat problems and has shortcuts. In that case, it will ignore the semantic information of the observation and miss the instructions of traffic lights, which may cause more vehicle collisions or traffic jams. To evaluate how much attention our framework pays to traffic lights, we count the percentage of each imitator passing the intersection while the traffic light is green in CARLA \emph{Nocrash} \emph{Dense}.

\begin{table}[h]
\caption{Percentage of obeying traffic lights}
\label{table:traffic light}
\centering
\begin{tabular}{l|*{3}{c}}
    \hline 
    Method & \textbf{BCOH} & \textbf{Keyframe} & \textbf{OURS} \\
    \hline
    Green light(\%)($\uparrow$) & 30.6 & 42.1  & \textbf{66.3} \\
    \hline
\end{tabular}
\end{table}

\begin{figure}[H]
    \centering
    \includegraphics[width=0.8\textwidth]{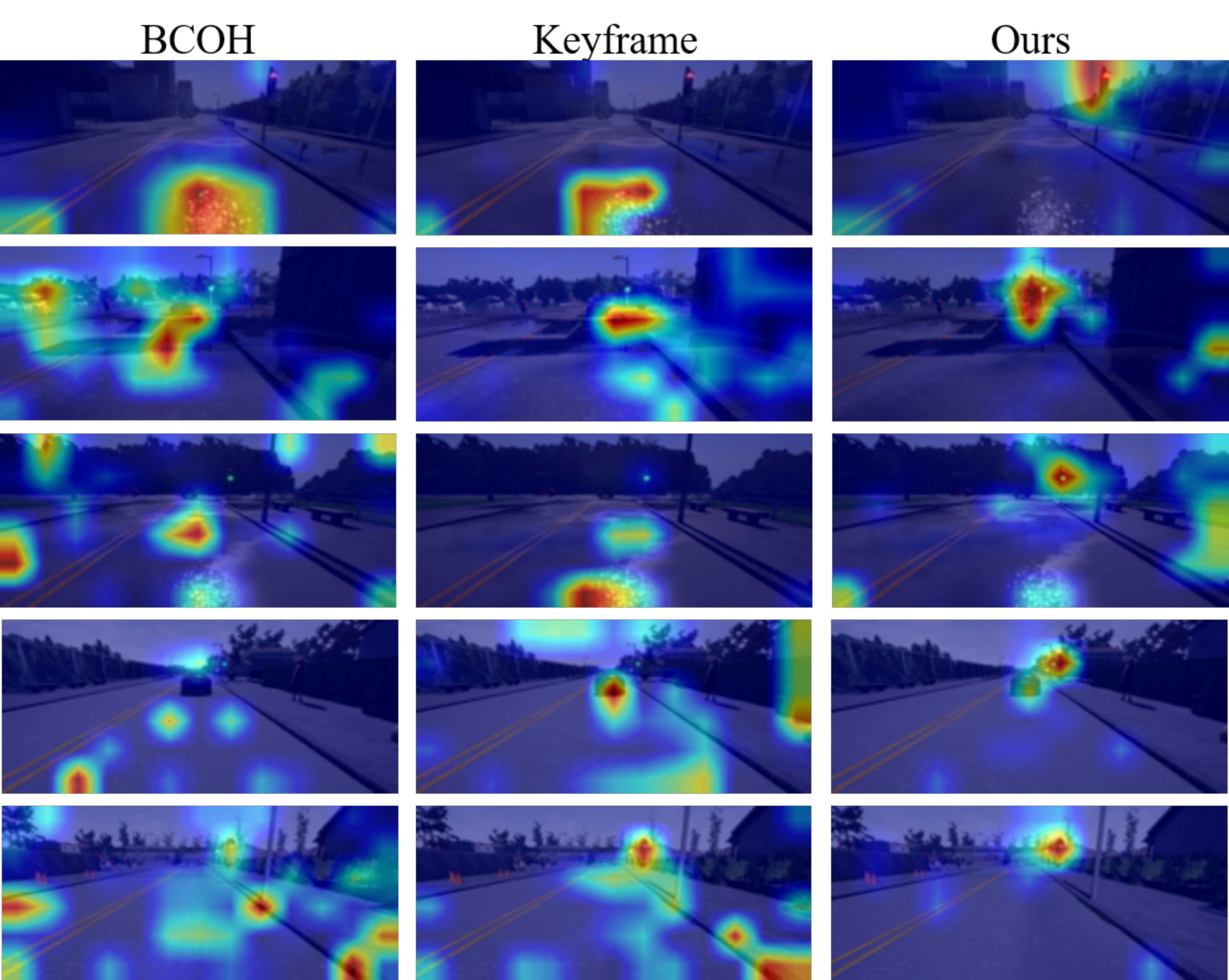}
    \caption{Attention maps generated by Grad-CAM \cite{2019gradcam}}
    \label{fig:red-light}
\end{figure}

Table~\ref{table:traffic light} shows the percentage of obeying traffic lights for all methods, and Fig.~\ref{fig:red-light} displays some visualization results about the observation with the traffic light in the validation set. Our method is the most compliant with traffic lights which helps our method achieve high \emph{\#SUCCESS}. The visualization results also show our method focuses more on the correct causal clue of the traffic light while BCOH and Keyframe concentrate on spurious road features.

\noindent \textbf{Minimize any previous action's impact~~} The model we propose only removes the information about last-step action $a_{t-1}$. However, the whole sequence can somehow have an impact on the shortcut learning of the predicted action $a_t$. In order to minimize any previous action's impact, we have done an interesting ablation by adding more objectives for the memory module. Intuitively, we define $m$ residual prediction branches for memory module, and the $i$th branch's objective is $a_t-a_{t-i}$. We tested it on CARLA \emph{NoCrash} Dense Benchmark. The success rate of one branch is 52.0\%. After increasing to 2 branches it slightly increases to 52.6\%; while further going to 4 branches degrade to 50.7\%. This suggests that having more branches can be beneficial, but having too many branches will not help. 

\noindent \textbf{The influence of two-streams architecture~~} To address the concern about the potential unfairness brought by the larger capacity of the two-stream network, we provide two extra ablations by running BCOH and KeyFrame with the two-stream architecture. We choose BCOH and KeyFrame since their performance is strong as shown in Table~\ref{tab:CARLA-results}. More specifically, we keep the two-stream architecture the same but replace the inputs to both streams as the observations with histories. We supervise the policy stream with the corresponding loss function of BCOH and KeyFrame. The results are shown in Table \ref{tab:two-streams}. Much lower \emph{\#SUCCESS} and higher \emph{\#TIMEOUT} of two-streams baselines indicate that two-streams architecture alone, without our method, suffers from severe copycat problems. We hypothesized that two stream architecture has even lower performance than their one stream counterparts because more parameters make it more vulnerable to the copycat problem. 

\begin{table}[h]
    \caption{Results of two-stream architecture on CARLA \emph{Nocrash Dense} benchmark}
    \label{tab:two-streams}
    \centering
    \begin{tabular}{c|c c}
       \hline
       Metrics & \emph{\#SUCCESS} & \emph{\#TIMEOUT}\\
       \hline
       Two-streams BCOH  & $23.7\pm3.1$ & $48.7\pm2.1$\\
       Two-streams Keyframe  & $38.7\pm2.5$ & $33.0\pm7.5$\\
       \hline
       BCOH  & $34.1\pm7.5$ & $36.1\pm14.5$\\
       Keyframe & $41.9\pm6.2$ & $24.8\pm7.9$\\
       OURS & $52.0\pm2.3$ &$23.3\pm7.6$\\ 
       \hline
    \end{tabular}
\end{table}

\section{Implementation Details of Experiments in MuJoCo-Image}

\begin{figure}[H]
    \centering
    \includegraphics[width=0.6\textwidth]{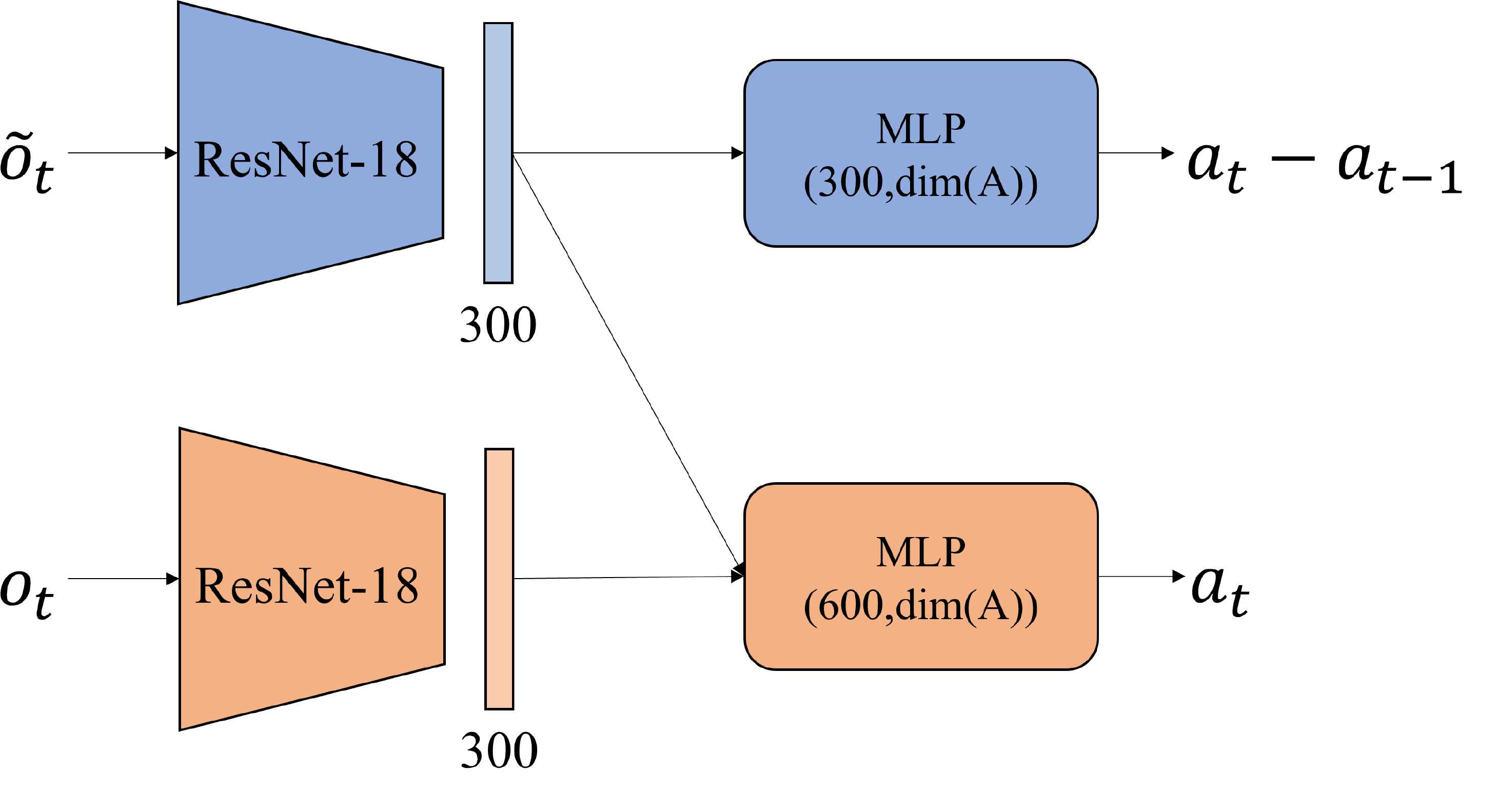}
    \caption{Our MuJoCo model: blue blocks are the memory extraction module; orange blocks are the policy module. dim(A) denotes the dimension of any action $a\in A$.}
    \label{fig:mujoco-detail}
\end{figure}

Fig.~\ref{fig:mujoco-detail} shows our model we used in MuJoCo. Both memory extraction module and policy module apply ResNet18 as their perception backbone to obtain a 300-dimensional feature and utilize this extracted feature to predict the defined objective via a one-layer MLP. The overall loss function is defined as follows
\begin{align}
    L_{\text{overall}} = L(a_t-a_{t-1}, a_t^{gt}-a_{t-1}^{gt}) + L(a_t, a_t^{gt}),
\end{align}
where all the symbols are the same with those in Eq.\eqref{eq: loss for ours}.\\

We apply the  hyper-parameters shown in Table~\ref{tab:hyper-params in mujoco}, including total training iterations, batch size, $\alpha$, loss function, optimizer setup, and other configurations about the learning rate (LR) scheduling, which has been explained in Sec.\ref{sec: other details}. 

\begin{table}[H]
    \caption{Hyper-parameters of experiments in MuJoCo-Image}
    \label{tab:hyper-params in mujoco}
    \centering
    \begin{tabular}{l|c}
         Configuration & Value\\
         \hline
         Total training iterations & 120k \\
         Batch size & 128 \\
         Loss function & $L_2$ \\
         \hline
         Optimizer & Adam \\
         Betas & (0.9, 0.999)\\
         Eps & 1e-08 \\
         Weight decay & 0.03 \\
         \hline
         Initial LR& 0.1 \\
         LR decay threshold & 40k \\
         LR decay rate & 0.1 \\
         Early Stop & True \\
         \hline
    \end{tabular}
\end{table}

\noindent \textbf{Other MuJoCo Environments~~} Following the original setting, we further conduct experiments in three more MuJoCo environments, including Ant, Reacher, and Humanoid. The demonstration trajectories are collected by TRPO experts. There are 1k samples for Ant, 5k samples for Reacher, and 200k samples for Humanoid, according to the task complexities. As shown in Table \ref{tab:new-mujoco}, our method outperforms the baselines in all these new MuJoCo Environments. We compare to BCOH and KeyFrame since they are the two stronger baselines as shown in Table~\ref{tab:MuJoCo-results}. 

\begin{table}[h]
    \caption{The average reward}
    \label{tab:new-mujoco}
    \centering
    \begin{tabular}{c|*{3}{c}}
       \hline
       Environment & Ant & Reacher & Humanoid \\
       \hline
       BCOH  & $746\pm96$ & $-81\pm8$ & $258\pm3$\\
       Keyframe & $790\pm85$  & $-71\pm5$ & $294\pm53$ \\
       OURS & $860\pm68$ & $-62\pm7$  & $372\pm20$\\
       \hline
    \end{tabular}
\end{table}

\end{document}